\documentclass[10pt,twocolumn,letterpaper]{article}

\usepackage{cvpr}              
\usepackage[accsupp]{axessibility}
\usepackage{graphicx}
\usepackage{amsmath}
\usepackage{amssymb}
\usepackage{booktabs}
\usepackage{color}
\usepackage[pagebackref,breaklinks,colorlinks]{hyperref}

\usepackage{colortbl}  
\usepackage{xcolor}
\usepackage{array}
\usepackage{hyperref}

\definecolor{hollywoodcerise}{rgb}{0.96, 0.0, 0.63}
\definecolor{lasallegreen}{rgb}{0.03, 0.47, 0.19}
\definecolor{hanpurple}{rgb}{0.32, 0.09, 0.98}
\definecolor{green(pigment)}{rgb}{0.0, 0.65, 0.31}

\hypersetup{colorlinks,linkcolor={red},citecolor={hollywoodcerise},urlcolor={red}}  

\usepackage[capitalize]{cleveref}
\crefname{section}{Sec.}{Secs.}
\Crefname{section}{Section}{Sections}
\Crefname{table}{Table}{Tables}
\crefname{table}{Tab.}{Tabs.}

\usepackage{multirow}

\def\cvprPaperID{7184} 
\def\confName{CVPR}
\def\confYear{2023}

\begin{document}

\title{Both Style and Distortion Matter: Dual-Path Unsupervised Domain Adaptation for Panoramic Semantic Segmentation}

\author{Xu Zheng$^{2}$ \quad Jinjing Zhu$^{1}$ \quad Yexin Liu$^{1}$ \quad Zidong Cao$^{1}$ \quad Chong Fu$^{2,4}$ \quad Lin Wang$^{1}$$^{,3}$\thanks{Corresponding author.}\\
$^{1}$AI Thrust, HKUST(GZ) \quad $^{2}$Northeastern University \quad $^{3}$Dept. of CSE, HKUST\\
$^{4}$Key Laboratory of Intelligent Computing in Medical Image, Ministry of Education, NEU, China\\\\
{\tt\small zhengxu128@gmail.com, zhujinjing.hkust@gmail.com, yliu292@connect.hkust-gz.edu.cn}\\
{\tt\small caozidong1996@gmail.com,  fuchong@mail.neu.edu.cn, linwang@ust.hk}
}
\maketitle
\begin{abstract}
\vspace{-4pt}
The ability of scene understanding has sparked active research for panoramic image semantic segmentation. However, the performance is hampered by distortion of the equirectangular projection (ERP) and a lack of pixel-wise annotations. For this reason, some works treat the ERP and pinhole images equally and transfer knowledge from the pinhole to ERP images via unsupervised domain adaptation (UDA). However, they fail to handle the domain gaps caused by: 1) the inherent differences between camera sensors and captured scenes; 2) the distinct image formats (\eg, ERP and pinhole images). In this paper, we propose a novel yet flexible dual-path UDA framework, DPPASS, taking ERP and tangent projection (TP) images as inputs. To reduce the domain gaps, we propose cross-projection and intra-projection training. The cross-projection training includes tangent-wise feature contrastive training and prediction consistency training.
That is, the former formulates the features with the same projection locations as positive examples and vice versa, for the models' awareness of distortion, while the latter ensures the consistency of cross-model predictions between the ERP and TP.
Moreover, adversarial intra-projection training is proposed to reduce the inherent gap, between the features of the pinhole images and those of the ERP and TP images, respectively. Importantly, the TP path can be freely removed after training, leading to no additional inference cost.
Extensive experiments on two benchmarks show that our DPPASS achieves +1.06$\%$ mIoU increment than the state-of-the-art approaches. \url{https://vlis2022.github.io/cvpr23/DPPASS}
\end{abstract}
\begin{figure}
    \centering
    \includegraphics[width=0.9\columnwidth]{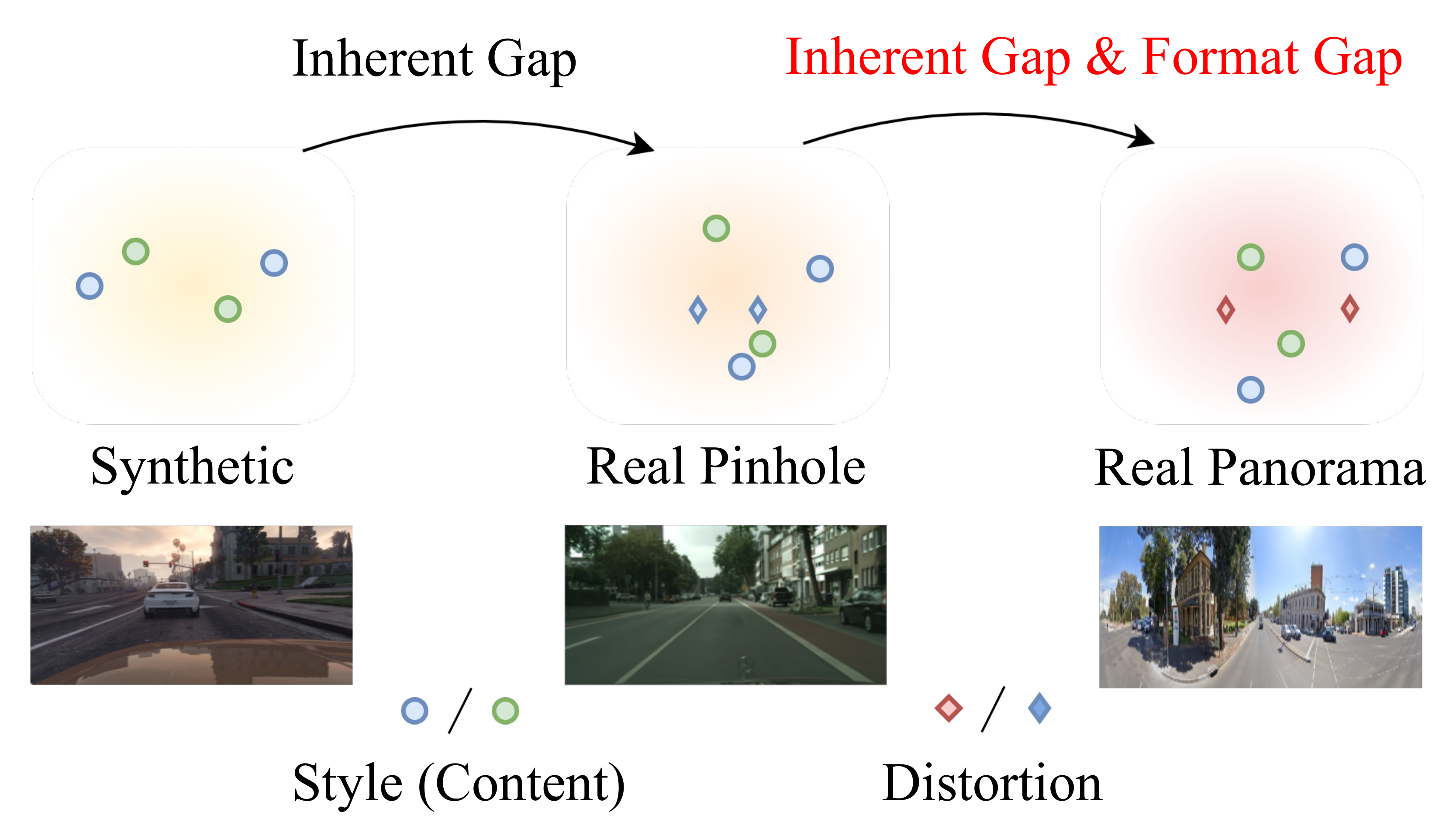}
    \caption{We tackle a new problem by addressing two types of domain gaps, \ie, the inherent gap (style) and format gap (distortion) between the pinhole and panoramic ($360^\circ$) images.
    } \vspace{-10pt}
\label{teaser}
\end{figure}

\vspace{-10pt}
\section{Introduction}
Increasing attention has been paid to the emerging $360^\circ$ cameras for their omnidirectional scene perception abilities with a broader field of view (FoV) than the traditional pinhole images~\cite{ai2022deep}. Intuitively, the ability to understand the surrounding environment from the panoramic images has triggered the research for semantic segmentation as it is pivotal to practical applications, such as autonomous driving~\cite{yang2021capturing,zhang2022bending} and augmented reality~\cite{schroeter2022inception}.
Equirectangular projection (ERP)~\cite{yoon2022spheresr} is the most commonly used projection type for the $360^\circ$ images \footnote{Here, panoramic and $360^\circ$ images are interchangeably used.} and can provide a complete view of the scene. However, the ERP type suffers from severe distortion in the polar regions, resulting in noticeable object deformation. This significantly degrades the performance of the pixel-wise dense prediction tasks, \eg, semantic segmentation. Some attempts have been made to design the convolution filters for feature extraction~\cite{zhang2022bending,tateno2018distortion,zhao2018distortion}; however, the specifically designed networks are less generalizable to other spherical image data. Moreover, labeled datasets are scarce, thus making it difficult to train effective $360^\circ$ image segmentation models. 

To tackle these issues, some methods,~\eg,~\cite{zhang2022bending} treat the ERP and pinhole images equally, like the basic UDA task, and directly alleviate the mismatch between ERP and pinhole images by adapting the neural networks trained in the pinhole domain to the $360^\circ$ domain via unsupervised domain adaptation (UDA). For instance, DensePASS~\cite{densepass} proposes a generic framework based on different variants of attention-augmented modules. Though these methods can relieve the need for the annotated $360^\circ$ image data~\cite{zhang2022bending}, they fail to handle the existing domain gaps caused by: 1) diverse camera sensors and captured scenes; 2) distinct image representation formats (ERP and pinhole images) and yield unsatisfied segmentation performance. Accordingly, we define these two types of domain gaps as the inherent gap and format gap (See Fig.\textcolor{red}{~\ref{teaser}}).

In this paper, we consider using the tangent projection (TP) along with the ERP. It has been shown that TP, the geometric projection~\cite{coxeter1961introduction} of the $360^\circ$ data, suffers from less distortion than the ERP. Moreover, the deep neural network (DNN) models designed for the pinhole images can be directly applied~\cite{eder2020tangent}. To this end, we propose a novel dual-path UDA framework, dubbed DPPASS, taking ERP and TP images as inputs to each path. The reason is that the ERP provides a holistic view while TP provides a patch-wise view of a given scene. For this, the pinhole images (source domain) are also transformed to the pseudo ERP and TP formats as inputs.
To the best of our knowledge, our work takes the first effort to leverage two projection formats, ERP and TP, to tackle the inherent and format gaps for panoramic image semantic segmentation. Importantly, the TP path can be freely removed after training, therefore, no extra inference cost is induced.

\begin{figure*}[t!]
    \centering
    \includegraphics[width=0.9\textwidth]{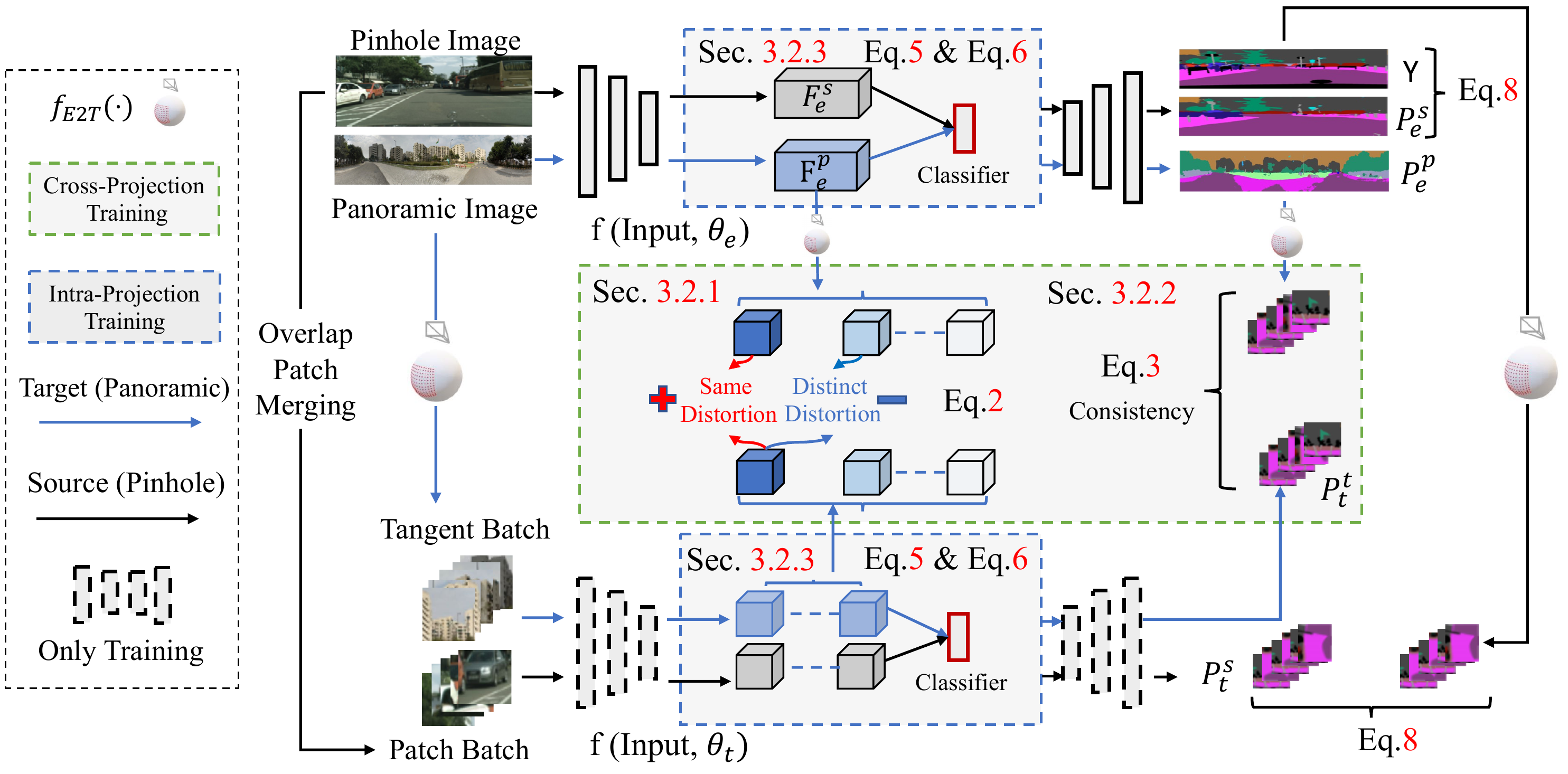}
    \vspace{-6pt}
    \caption{Overview of the proposed DPPASS framework, consisting of two models  $f(Input,\theta_e)$, $f(Input,\theta_t)$. 
    Our method has two major components: cross- and intra- projection training. The cross-projection training explores prediction consistency training and tangent-wise feature contrastive training, and the intra-projection training employs adversarial training for transferring knowledge from the pinhole images to panoramas.}
    \vspace{-12pt}
    \label{framework}
\end{figure*}

Specifically, as shown in Fig.\textcolor{red}{~\ref{framework}}, the cross-projection training is proposed at both the feature and prediction levels for tackling the challenging format gap (Sec.~\ref{Cross-Projection}). At the feature level, the tangent-wise feature contrastive training aims at mimicking the tangent-wise features with the same distortion and discerning the features with distinct distortion, to further learn distortion-aware models and decrease the format gap.
Meanwhile, the less distorted tangent images are used in the prediction consistency training. It ensures the consistency between the TP predictions and the tangent projections of the ERP predictions for models' awareness of the distortion variations. 
For the long-existing inherent gap, the intra-projection training imposes the style and content similarities between the features from the source and target domains for both the ERP and TP images (Sec.~\ref{Intra-Projection}).
As such, we can reduce the large inherent and format gaps between the $360^\circ$ and pinhole images by taking advantage of dual projections. 

We conduct extensive experiments from the pinhole dataset, Cityscapes~\cite{Cityscapes}, to two $360^\circ$ datasets: DensePASS~\cite{densepass} and WildPASS~\cite{wildpass}. The experimental results show that our framework surpasses the existing SOTA methods by 1.06$\%$ on the DensePASS test set. 
In summary, our main contributions are summarized as follows: (I) We study a new problem by re-defining the domain gaps between $360^\circ$ images and pinhole images as two types: the inherent gap and format gap. (II) We propose the first UDA framework taking ERP and tangent images to reduce the types of domain gaps for semantic segmentation. (III) We propose corss- and intra- projection training that take the ERP and TP at the prediction and feature levels to reduce the domain gaps.

\section{Related work}
\noindent\textbf{Panoramic image semantic segmentation} 
Most existing works~\cite{swin,PVT, zheng2022transformer} on semantic segmentation focused on pinhole images having a limited FoV; consequently, the performance significantly drops when they are applied to the $360^\circ$ images. The panoramic image segmentation has to tackle two challenges: 1) the inevitable distortion and object deformation in the ERP images and 2) a lack of accurately labeled data~\cite{wang2014automatic,varga2017super,narioka2018understanding,deng2019restricted}.

In the literature, the mainstream methods can be divided into three types: supervised learning methods ~\cite{yang2019pass,su2019kernel}, UDA methods ~\cite{densepass,gu2021pit,p2pda}, and unsupervised contrastive learning methods~\cite{jaus2021panoramic}. Among the supervised learning methods, ~\cite{tateno2018distortion,jiang2019spherical,zhang2019orientation} focus on designing distortion-aware and trainable deformable convolution layers for dense depth and semantic prediction on the panoramic images.~\cite{yang2019pass,yang2020ds,wildpass}, on the other hand, explore the multi-source learning schemes to train the network on the pinhole images and deploy it to the unseen panoramas. The unsupervised contrastive learning methods learn robust feature representations, allowing the network model to better generalize to data from a different distribution~\cite{jaus2021panoramic}. 

As labeled panoramic image data is limited, UDA methods ~\cite{densepass,p2pda} have been proposed to transfer knowledge from the output, and feature space of pinhole images to those of the panoramic images. 
However, these methods do not fully consider the domain gaps between the pinhole and $360^\circ$ images. Also, the domain gap between different projection types, \eg, ERP and TP, of $360^\circ$ images has been neglected. For this reason, we formulate the two types of domain gaps between the $360^\circ$ and pinhole images: 1) inherent gaps caused by the different sensors and scenes and 2) format gap caused by the difference of image representation formats. In our work, we are the first to tackle both domain gaps simultaneously.

\noindent\textbf{Unsupervised domain adaptation.} UDA takes the labeled source domain data and unlabeled target domain data as inputs and trains the network model in an unsupervised manner to enhance the generalization capacity to the targeted domain. The UDA techniques are critical for semantic segmentation, especially when labeled data is particularly scarce. The mainstream UDA methods rely on self-training with pseudo labels ~\cite{guo2021metacorrection,zhang2017curriculum}, adversarial learning ~\cite{chang2019all,hoffman2018cycada,cheng2021dual,touvron2021training}, entropy minimization ~\cite{du2019ssf,toldo2021unsupervised}, and self-ensembling ~\cite{french2017self,ioffe2015batch,perone2019unsupervised}.  The self-training methods create pseudo labels for the target data and gradually adapt through the iterative improvement ~\cite{tarvainen2017mean}.  However, pseudo labels are often error-prone. To mitigate this problem, ~\cite{zou2019confidence,shin2020two,tranheden2021dacs} attempt to make self-training less sensitive to the incorrect pseudo labels.  The adversarial learning methods mainly adopt generative adversarial networks (GANs) to learn a shared latent representation between the two domains while maintaining the domain-specific characteristics. The entropy minimization methods aim to enforce structural consistency across domains by applying it jointly with the square losses~\cite{chen2019domain} or adversarial loss ~\cite{vu2019advent}. 
Our work explores the potential of adversarial learning and ensemble learning to tackle the two types of domain gaps between the pinhole and $360^\circ$ images, mentioned in the introduction. We focus on exploiting the feature embeddings and predictions from the ERP and TP paths to transfer knowledge from the pinhole image domain to the 360$^\circ$ image domain.

\section{Methodology}
\subsection{Overview}
An overview of the proposed DPPASS is depicted in Fig. ~\ref{framework}.
Given the target domain data consisting a set of $n$ unlabeled ERP $P$ = {$E_p^1$, ..., $E_p^n$} and a set of annotated pinhole images in the source domain are transformed to the $m$ pseudo ERP $S$ = {($I_s^1$, $Y_s^1$), ..., ($I_s^m$, $Y_s^m$)}. The tangent images $T$ = {$E_t^1$, ..., $E_t^{18n}$} are projected by function $f_{E2T}(\cdot)$ (see Fig.~\ref{tangent}) from the ERP image set $P$, and the pseudo tangent image set $T^*$ = {$I_{t^*}^1$, ..., $I_{t^*}^{18m}$} are projected by the overlap patch merging from the pseudo ERP set $S$. 
$E_p^i$ is the $i$-th ERP image with spatial dimensions $H \times W$, $I_s^i$ is the $i$-th pseudo ERP (pinhole) image from the source domain has the same spatial dimensions as $E_p^i$ with its corresponding pixel-level label $Y_s^i$ $\in$ $(1, C)^{H \times W}$, where $C$ is the number of classes. $I_t^i$ and $I_{t^*}^i$ are the tangent and pseudo tangent images projected from $E_p^i$ and $I_s^i$.
We propose a novel dual-projection UDA framework to minimize the two domain gaps, namely the inherent gap and format gap, simultaneously for panoramic image semantic segmentation. The two network models $f(Input,\theta_e)$ and $f(Input,\theta_t)$ in the framework are based on the vision transformers~\cite{segformer}. $f(Input,\theta_e)$ takes an ERP image $E_p^i$ $\in$ $P$ and the pseudo ERP $I_s^i$ $\in$ $S$ as the input, and $f(Input,\theta_t)$ takes the tangent images $I_t^i$ and the pseudo tangent images $I_{t_*}^i$ as inputs:
\begin{equation}
\setlength{\abovedisplayskip}{3pt}
\setlength{\belowdisplayskip}{3pt}
 \begin{aligned}
   P_e^p, F_e^p = f(E_p^i,\theta_e), P_e^s, F_e^s = f(I_s^i,\theta_e), \\
   P_t^t, F_t^t = f(E_t^i,\theta_t), P_t^s, F_t^s = f(I_{t_*}^i,\theta_t),   
\end{aligned}  
\end{equation}
where $P_e^p$ and $P_e^s$ are the ERP format predictions, $P_t^t$ and $P_t^s$ are the tangent format predictions, $F_e^p$, $F_e^s$, $F_t^t$ and $F_t^s$ are the high-level features. The model $f(Input,\theta_e)$ in the ERP path and the model $f(Input,\theta_t)$ in the TP path are trained and optimized individually. Importantly,  the TP path $f(Input,\theta_t)$ can be removed after training, and only $f(Input,\theta_e)$ is used for inference, leading to no additional inference cost.

Based on the aforementioned definitions for the two types of domain gaps, our proposed framework consists of two key components. Firstly, cross-projection training is proposed in both the prediction and feature spaces. As the network models in two paths exploit ERP and TP images as the representations of the 360$^{\circ}$ data, $f(P,\theta_e)$ and $f(T,\theta_t)$ can learn the inherent gap and format gap together from the patch-wise view of TP images and the holistic view of ERP images based on the feature and prediction perspectives. 
Secondly, the intra-projection training is designed in each path to tackle the inherent domain gap. 

In the cross-projection training, the prediction consistency training is proposed to fully utilize the distinct characteristics of tangent images to reduce the format gap. Because the tangent images have less distortion and object deformation, we first leverage the geometric projection function  $f_{E2T}(\cdot)$ to transform the ERP image to the TP patches. Then, the consistency regularization loss is employed to ensure the consistency of the cross-model predictions. 
For the feature-level cross-projection training, tangent-wise feature contrastive training (TFCT) is proposed to align the tangent-wise high-level features to reduce the format gap. We now describe these components in detail 

\subsection{Cross-Projection Training}
\label{Cross-Projection}
\subsubsection{Tangent-wise feature contrastive training }
\begin{figure}
    \centering
    \includegraphics[width=0.42\textwidth]{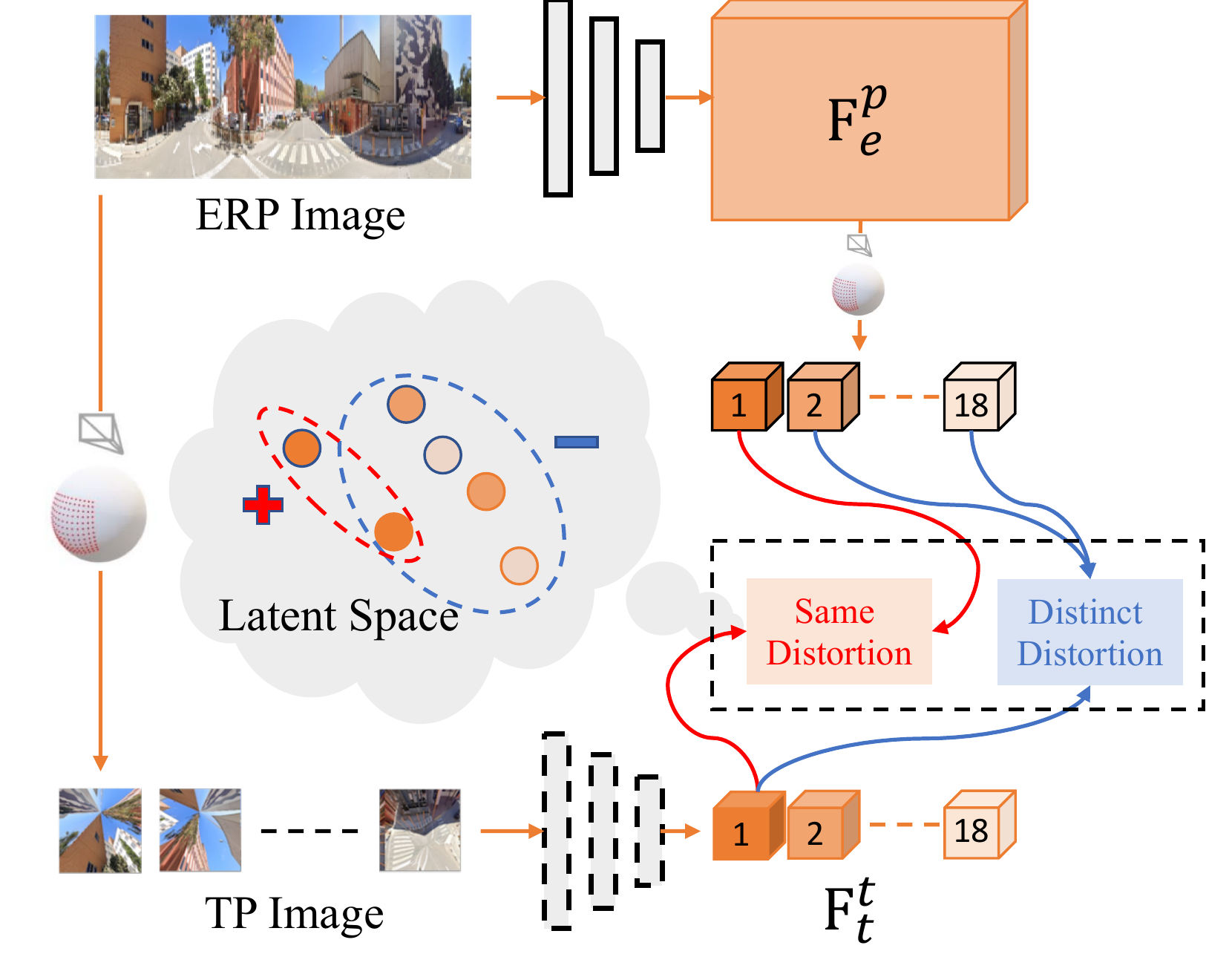}
    \caption{Overview of the proposed the tangent-wise feature contrastive training (TFCT) module.}
    \vspace{-12pt}
    \label{TFCT}
\end{figure}
We explore to impose consistency between the ERP and TP paths at the feature map level to diminish the format gap. As the models in the ERP path and TP path take the ERP and tangent images as inputs, respectively, the extracted high-level features are heterogeneous to each other (see Fig.~\ref{TFCT}). To align the features from the dual projection paths, we propose to process the features $F_e^p$ from $f(P,\theta_e)$ by $f_{E2T}(\cdot)$ to match the features $F_t^t$ extracted from $f(T,\theta_t)$. After aligning the features $F_t^t = {F_{t1}^t, F_{t2}^t, ..., F_{t18}^t}$ and $f_{E2T}(F_e^p) = {F_{e1}^p, F_{e2}^p, ..., F_{e18}^p}$ from two models, a contrastive learning strategy is applied to reduce the format gaps caused by the distortion. This is because tangent images are the geometric projection of the 360$^\circ$ image data; thus the distortion at different locations is different. 

For this reason, as shown in Fig.~\ref{TFCT}, we divide the features according to the TP $f_{E2T}(\cdot)$, and then align them by the projection locations. Specifically, the TP images, which are oriented in the same position and angle, have the same distortion and object deformation. 
That is, in Fig.~\ref{TFCT}, for the feature representation (cube), only the ones in the same color having the same projection locations are the positive example pairs, and the other representations in this batch are all negative example pairs.
Consequently, our proposed TFCT module aims to maximize the consistency between the positive pairs, \eg, $F_{t1}^t$ and $F_{e1}^p$, which represent the same form of distortion of the TP format. By contrast, the tangent features extracted from tangent images at different projection locations, are formulated as negative pairs, \eg, $F_{t1}^t$ and $F_{e2}^p$. 
Given the two feature sequences $F_t^t = {F_{t1}^t, F_{t2}^t, ..., F_{t18}^t}$ and $f_{E2T}(F_e^p) = {F_{e1}^p, F_{e2}^p, ..., F_{e18}^p}$, we formulate the contrastive training loss, based on the InfoNCE~\cite{oord2018representation}, which is:
\begin{equation}
\setlength{\abovedisplayskip}{3pt}
\setlength{\belowdisplayskip}{3pt}
L_{fc} = \frac{1}{F_i} \sum_{f_+ \in F_i}  -log \frac{exp(f_+ / \tau)}{exp(f_+ / \tau) + \sum_{f}exp(f_{-} / \tau)  },
\label{loss_tfct}
\end{equation}
where $f_+$ denotes the positive examples which stand for the same position in ERP, \eg, $F_{t1}^t$ and $F_{e1}^p$, the negative examples $f_{-}$ denote the tangent-wise features extracted from different locations in ERP, \eg, $F_{t1}^t$ and $F_{e2}^p$, $F_i$ denotes all the tangent-wise features in one batch and the $\tau$ is the temperature hyper-parameter.

Previous methods~\eg,~\cite{jaus2021panoramic}, applying contrastive learning to panoramic semantic segmentation, have to maintain a large memory bank to store the negative examples. This leads to a large capacity of memory and high training costs. Our TFCT module takes the tangent images projected from distinct locations in the same training batch as the negative examples; therefore, it requires less memory during the training process.
\subsubsection{Prediction consistency training}
We present the prediction consistency training to address the format gap which is mainly caused by the distortion of the ERP images. ERP is a common spherical image representation format based on the simple relation between rectangular and spherical coordinates, making it suffer from severe image distortion and object deformation. These inevitable problems impede applying UDA to the pinhole and panoramic image domains. 

\begin{figure}[t!]
    \centering
    \includegraphics[width=0.35\textwidth]{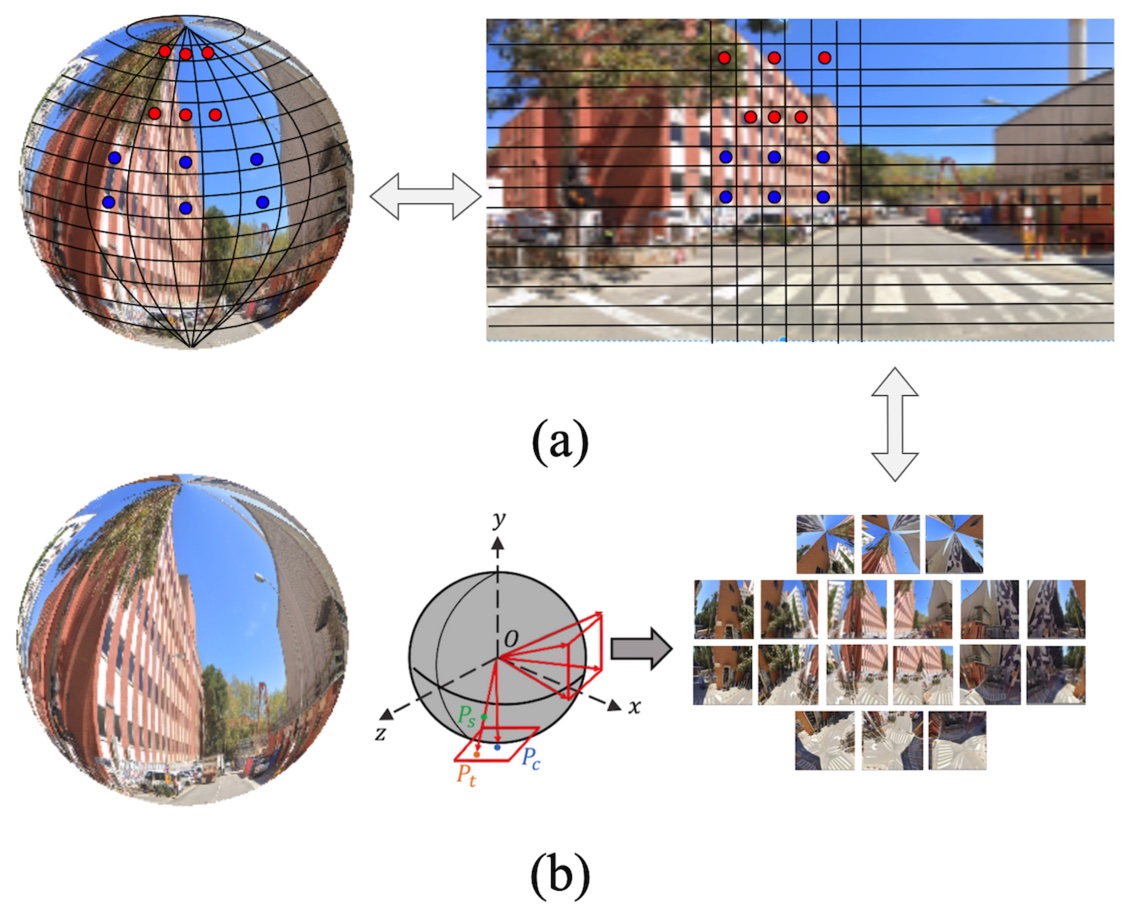}
    \caption{(a) Visualization of distortion. (b) Tangent image facilitates the transferable and scalable panoramic image representation. We use 18 tangent images to project the ERP as in \cite{OmniFusion}.}
    \vspace{-12pt}
    \label{tangent}
\end{figure}

Recently, TP~\cite{eder2020tangent} is shown to better mitigate the distortion, and thus the deep learning models developed for the pinhole images can be directly applied to the TP images. 
Though TP has less distortion than the ERP, ERP has a more holistic awareness of the surrounding scene. Accordingly, we leverage ERP and tangent images together to bridge the domain gaps caused by the distortion. Specifically, as depicted in Fig.~\ref{tangent}, the network model $f(Input,\theta_e)$ in the ERP path predicts a semantic label from an ERP image input while the model $f(Input,\theta_t)$ in the TP path processes the tangent image patches. 
We then project the ERP prediction $P_e^p$ with $f_{E2T}(\cdot)$ to get patch prediction maps $f_{E2T}(P_e^p)$ of the TP format. 
The consistency regularization is finally applied between $P_t^t$ and $f_{E2T}(P_e^p)$ to make the network models in the dual paths aware of the distortion discrepancies between the ERP and TP images. For convenience, the consistency regularization loss is formulated by the KL-Divergence:
\begin{equation}
\setlength{\abovedisplayskip}{3pt}
\setlength{\belowdisplayskip}{3pt}
\label{loss_con}
        \mathcal{L}_{pc} = \sum_{i=1}^{18} f_{E2T}(P_{ei}^p)\log \frac{f_{E2T}(P_{ei}^p))}{P_{ti}^t},
\end{equation}
where $P_{ei}^p$ and $P_{ti}^t$ denote the $i$th tangent-wise predictions.
The distinct representations of the same sphere data have the same semantic and content information, thus leading to better dual models' awareness of the distortion differences between ERP and TP. The tangent projection also can be treated as a data augmentation approach, which is thus applied before and after the model forward prediction for the consistency training.

\subsection{Intra-Projection Training}
\label{Intra-Projection}
We propose intra-projection training to decrease the inherent domain gap between panoramic images and pinhole images in each projection path. The main goal is to regularize the learning of the internal features from source and target domains, which are extracted from the same model and in the same format (\ie, ERP and TP). Specifically, a domain classifier $f(F, \theta_d)$ is added after the feature extractor to distinguish the features extracted from panoramas (TP images) or pinhole images (pseudo TP images). 

We denote $d$ as the binary variable for the extracted features $F$, which indicates whether $F$ is extracted from the panoramas ($F_e^p$ and $F_e^s$ if $d_t$ = 1) or from the pinhole images ($F_t^t$ and $F_t^s$ if $d_s$ = 0). The classifier and the feature extractor (encoder) are optimized individually, the classifier is trained to better distinguish the features extracted from different domains while the feature extractor is optimized to generate domain-invariant features:
\begin{equation}
\setlength{\abovedisplayskip}{3pt}
\setlength{\belowdisplayskip}{3pt}
    D_u = f(F_e^p, \theta_d), D_s = f(F_e^s, \theta_d),
\end{equation}
where the $D_t$ and $D_s$ are the classification predictions of the features.
For training the classifier, we aim to minimize the supervised loss with Binary Cross Entropy (BCE) as:
\begin{equation}
\setlength{\abovedisplayskip}{3pt}
\setlength{\belowdisplayskip}{3pt}
\begin{aligned}
     L_d^c = - [(d_t \cdot log(D_t) + (1 - d_t) \cdot log(1 - D_t)) \\
     + (d_s \cdot log(D_s) + (1 - d_s) \cdot log(1 - D_s))].
\end{aligned}
\label{gan3}
\end{equation}
To reduce the inherent domain gap, the domain adaptation loss for the feature extractor (encoder) is measured by BCE as follows:
\begin{equation}
\setlength{\abovedisplayskip}{3pt}
\setlength{\belowdisplayskip}{3pt}
\begin{aligned}
     L_d = - [(d_t \cdot log(D_s) + (1 - d_t) \cdot log(1 - D_s)) \\
     + (d_s \cdot log(D_t) + (1 - d_s) \cdot log(1 - D_t))].
\end{aligned}
\label{gan4}
\end{equation}
In summary, through the dual-projection regularization, we make the two models, $f(P,\theta_e)$ and $f(T,\theta_t)$, learn the domain gaps between pinhole and panoramic images at ERP and tangent image scales.

\subsection{Optimization}
The training objective containing four losses is defined as:
\begin{equation}
\setlength{\abovedisplayskip}{3pt}
\setlength{\belowdisplayskip}{3pt}
    \mathcal{L} = \mathcal{L}_s + \alpha \cdot \mathcal{L}_d + \beta \cdot \mathcal{L}_{pc} + \mathcal{L}_{fc},
\end{equation}
where the $L_s$ is the supervised loss on the Cityscapes dataset, the $L_g$ refers to the intra-projection loss, $L_{pc}$ denotes the prediction consistency loss, the $L_{fc}$ is the tangent-wise feature contrastive loss between $f(Input,\theta_e)$ and $f(Input,\theta_t)$ and the $\alpha$ and $\beta$ are the trade-off weight of the proposed loss terms. The supervised loss on Cityscapes is formulated using the standard Cross-Entropy (CE) loss:
\begin{equation}
\setlength{\abovedisplayskip}{3pt}
\setlength{\belowdisplayskip}{3pt}
    \mathcal{L}_s = -\sum_{i=0}^{C} Y_ilog(P_{e}^s). 
\end{equation}
Especially, for the network model in the TP path, the pseudo tangent images are obtained through random crops on the Cityscapes train set.

\begin{figure*}[t!]
    \centering
    \includegraphics[width=\textwidth,height=125pt]{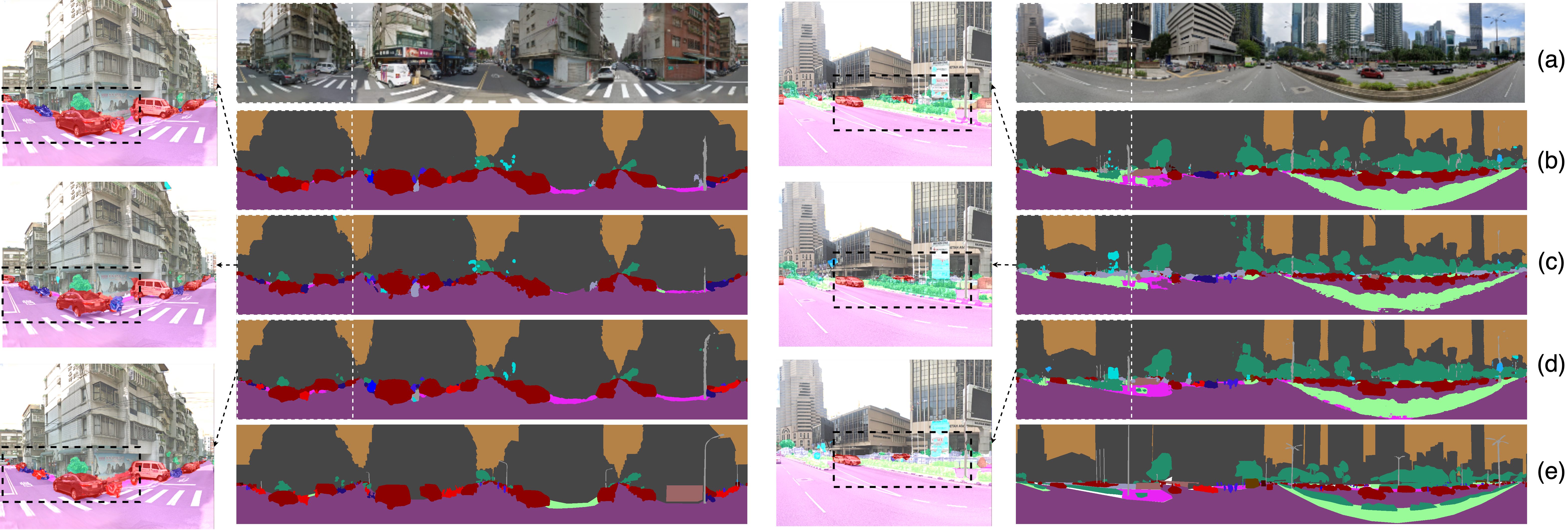}
    \caption{Example visualization results from DensePASS test set. (a) Input, (b) Fully supervised Segformer-B1 without domain adaptation~\cite{segformer}, (c) Trans4PASS-T~\cite{zhang2022bending}, (d) DPPASS-T, and (e) Ground truth.}
    \label{360results}
\end{figure*}

\begin{table}[t!]
\centering
\small
\resizebox{0.49\textwidth}{!}{
\begin{tabular}{ccccl}
\toprule
Network & Backbone      & CS   & DP   & GAPs \\ \midrule

 & ResNet-50     & 78.6 & 29.5 & 49.1 \\ 
\multirow{-2}{*}{PSPNet~\cite{PSPNet}}       & ResNet-101    & 79.8 & 30.4 & 49.4 \\ \midrule
                                                      & ResNet-50     & 80.1 & 29.0 & 51.1 \\ 
\multirow{-2}{*}{DeepLabv3+~\cite{Deeplabv3plus}}                           & ResNet-101    & 80.9 & 32.5 & 48.4 \\ \midrule
& ResNet-50     & 74.5 & 29.9 & 44.6 \\ 
& ResNet-101    & 75.8 & 28.8 & 47.0 \\ 

\multirow{-3}{*}{Semantic-FPN~\cite{fpn}} & PVT-T         & 71.5 & 31.2 & 40.3 \\ \midrule
SETR ~\cite{SETR}                                                  & Transformer-L & 77.9 & 36.1 & 41.8 \\ \midrule                             
& Mit-B1        & 78.5 & 38.5 & 40.0 \\ 
\multirow{-2}{*}{Segformer~\cite{segformer}}    & MiT-B2        & 81.0 & 42.4 & 38.6 \\ \midrule
ERFNet~\cite{erfnet}                                                & ERFNet        & 72.1 & 16.7 & 55.4 \\ 
FANet~\cite{FANet}                                                 & ResNet-34     & 71.3 & 26.9 & 44.4 \\ 
DANet~\cite{DANet}                                                 & ResNet-50     & 79.3 & 28.5 & 50.8 \\ 
                                     & ERFNet        & 72.1 & 34.1 & 38.0 \\ 
P2PDA ~\cite{p2pda}                                       & ResNet-34     & 71.3 & 33.1 & 38.2 \\ 
                                      & ResNet-50     & 79.3 & 39.8 & 39.5 \\ \midrule & Trans4PASS-T  & 79.1 &41.5 & 37.6 \\ 
\multirow{-2}{*}{Tarns4PASS~\cite{zhang2022bending}}                        & Trans4PASS-S  & 81.1 & 44.8 & 36.3 \\ \midrule
& ResNet-34        & 75.4 & 38.9   & 36.5 \\
& ResNet-50        & 78.6 & 42.3   & 36.3 \\
& Mit-B1        & 76.3 & 42.4   & 36.1      \\ 
\multirow{-4}{*}{DPPASS(Ours)}         & Mit-B2        & 80.1 & \textbf{48.6} & \textbf{32.4}     \\ \bottomrule
\end{tabular}}
\caption{Performance gaps of semantic segmentation methods from Cityscapes dataset (CS) to DensePASS dataset (DP).}
\vspace{-6pt}
\label{domaingaps}
\end{table}

\begin{table*}[t!]
\centering
\resizebox{\textwidth}{!}{
\begin{tabular}{ccccccccccccccccccccc}
\toprule
Method             & mIoU  & \rotatebox{90}{road}  & \rotatebox{90}{sidewalk} & \rotatebox{90}{building} & \rotatebox{90}{wall}  & \rotatebox{90}{fense} & \rotatebox{90}{pole}  & \rotatebox{90}{traffic Light} & \rotatebox{90}{traffic Sign} & \rotatebox{90}{tegetation} & \rotatebox{90}{terrain} & \rotatebox{90}{sky}   & \rotatebox{90}{Person} & \rotatebox{90}{rider} & \rotatebox{90}{car}   & \rotatebox{90}{truck} & \rotatebox{90}{bus}   & \rotatebox{90}{train} & \rotatebox{90}{motorcycle} & \rotatebox{90}{bicycle} \\ \midrule
ERFNet& 16.65 & 63.59 & 18.22    & 47.01   & 9.45 & 12.79 & 17.00  & 8.12  & 6.41 & 34.24 & 10.15 & 18.43 & 4.96  & 2.31  & 46.03 & 3.19  & 0.59  & 0.00  &8.30  & 5.55   \\
PASS(ERFNet)      & 23.66 & 67.84 & 28.75    & 59.69    & 19.96 & 29.41 & 8.26  & 4.54          & 8.07         & 64.96      & 13.75   & 33.50 & 12.87  & 3.17  & 48.26 & 2.17  & 0.82  & 0.29  & 23.76      & 19.46   \\ 
Omni-sup(ECANet)   & 43.02 & 81.60 & 19.46    & 81.00    & 32.02 & 39.47 & 25.54 & 3.85          & 17.38        & 79.01      & 39.75   & 94.60 & 46.39  & 12.98 & 81.96 & 49.25 & 28.29 & 0.00  & 55.36      & 29.47   \\ 
P2PDA(Adversarial) & 41.99 & 70.21 & 30.24    & 78.44    & 26.72 & 28.44 & 14.02 & 11.67         & 5.79         & 68.54      & 38.20   & 85.97 & 28.14  & 0.00  & 70.36 & 60.49 & 38.90 & 77.80 & 39.85      & 24.02   \\ 
PCS& 53.83 & 78.10 & 46.24 & 86.24  & 30.33 &45.78 & 34.04  & 22.74  & 13.00 & 79.98 & 33.07 & 93.44 & 47.69  & 22.53  & 79.20 & 61.59  & 67.09  & 83.26  & 58.68  & 39.80   \\
Trans4PASS-T $\dagger$ & 53.18 & 78.13 & 41.19    & 85.93    & 29.88 & 37.02  & 32.54 & 21.59         & 18.94        & 78.67      &45.20   &93.88 & 48.54  & 16.91 & 79.58 & 65.33 & 55.76 & 84.63 & 59.05      & 37.61   \\ 
Trans4PASS-S $\dagger$ & 55.22 & 78.38 & 41.58    & 86.48 & 31.54 & 45.54  & 33.92 & 22.96 & 18.27 & 79.40 & 41.07 & 93.82 & 48.85  & 23.36 &81.02 &67.31 &69.53 & 86.13 & 60.85 & 39.09   \\ 
\rowcolor{gray!10} DPPASS-T(Ours)  &\textbf{55.30} &78.74 &46.29 &87.47 &\textbf{48.62} &40.47 &\textbf{35.38} &24.97 &17.39 &79.23 &40.85 &93.49 &52.09 &\textbf{29.40} &79.19 &58.73 &47.24 &\textbf{86.48} &66.60 &38.11 \\ 
\rowcolor{gray!20} DPPASS-S(Ours) &\textbf{56.28} &78.99 &\textbf{48.14} &\textbf{87.63} &42.12 &44.85 &34.95 &\textbf{27.38} &\textbf{19.21} &78.55 &43.08 &92.83 &\textbf{55.99} &29.10 &80.95 &61.42 &55.68 &79.70 &\textbf{70.42} &38.40         \\ 
\bottomrule
\end{tabular}}
\vspace{-6pt}
\caption{Per-class results of the SOTA panoramic image semantic segmentation methods on DensePASS test set. }
\vspace{-6pt}
\label{perclass}
\end{table*}

\begin{table}[]
\small
\resizebox{0.49\textwidth}{!}{
\begin{tabular}{ccc}
\toprule
Method                      & Backbone           & mIoU(\%) \\ \midrule
\multirow{2}{*}{Source domain Supervised} & Segformer-B1       & 47.90         \\ 
                            & Segformer-B2       &  54.11       \\ \midrule
Trans4PASS-T~\cite{zhang2022bending}                & Segformer-B1 &54.67          \\
Trans4PASS-S~\cite{zhang2022bending}                & Segformer-B2 &62.91          \\ 
\rowcolor{gray!10} DPPASS-T(Ours)                   & Segformer-B1       & \textbf{60.38}          \\ 
\rowcolor{gray!20} DPPASS-S(Ours)                   & Segformer-B2       & \textbf{63.53}          \\ \bottomrule
\end{tabular}}
\caption{Experimental results of the SOTA panoramic image semantic segmentation methods on WildPASS test set.}
\vspace{-6pt}
\label{wildpass}
\end{table}

\section{Experiments}
To evaluate the performance of our method, we conduct extensive experiments on two benchmark datasets including DensePASS~\cite{densepass} and WildPASS~\cite{wildpass}. Experimental results demonstrate the superiority of our proposed DPPASS.
\subsection{Datasets and Implementation Details}
\noindent{\textbf{Cityscapes~\cite{Cityscapes}}} is an autonomous driving dataset that contains urban street scenes recorded from 50 different cities with precise pixel-wise annotations of 19 semantic categories. 
\noindent{\textbf{DensePASS~\cite{densepass}}} is a panoramic dataset and contains 2,000 images for training and 100 precise annotated images for testing. 
\noindent{\textbf{WildPASS~\cite{wildpass}}} is a panoramic dataset designed to capture diverse scenes from all around the globe and contains 2500 panoramas. 

\noindent{\textbf{Evaluation.}} We take the mean Intersection-over-Union (mIoU) as the evaluation metric in both the source and target domains. Our framework is evaluated on the DensePASS / WildPASS validation set via test at a single scale, and the resolution is 400 $\times$ 2048. 

\noindent{\textbf{Implementation details.}} 
Our framework is implemented with Pytorch and trained on multiple NVIDIA GPUs. Both models in our framework are based on the efficient Segformer~\cite{segformer}. Ours-T and Ours-S are two implementations of our framework, which are based on SeformerB1 and SegormerB2, respectively. 

\subsection{Inevitable Domain Gaps}
As shown in Tab.~\ref{domaingaps}, there are large segmentation performance drops from Cityscapes to DensePASS datasets. 
Even though some recent high-performance transformer-based networks~\cite{segformer,SETR} have better results on the pinhole images (Cityscapes) than the convolutional neural networks (CNNs), the performance on the panoramic images is still unsatisfying. Meanwhile, although some well-designed distortion-aware frameworks, \eg, ~\cite{zhang2022bending}, have been proposed for panoramic semantic segmentation, the domain gaps and performance drops are still large. 

Our proposed DPPASS, empowered by the unified vision transformer backbone MiT~\cite{segformer} without using the deformable components, achieves better performance than the SOTA methods. The reported results of our DPPASS on DensePASS are the average predictions of the two models. With Mit-B2 backbone, our method achieves 48.6$\%$ mIoU on DensePASS test set and the performance drop is only 32.4$\%$, which is 3.9$\%$ mIoU better than the SOTA method Trans4PASS (32.4$\%$ vs. 36.3$\%$). This indicates that our method effectively reduces the inherent gap and format gap by exploring the knowledge from ERP and TP paths, thus yielding better segmentation performance.

\subsection{Experimental Results}
We first train our DPPASS with the DenPASS train set and Cityscapes train set and then evaluate with the DensePASS test set. Tab.~\ref{perclass} shows the quantitative results. We compare our framework with some SOTA panoramic segmentation methods: PASS~\cite{yang2019pass}, Omni-sup~\cite{yang2020omnisupervised} and Trans4PASS~\cite{zhang2022bending}; domain adaptation approaches: P2PDA~\cite{p2pda} and PCS~\cite{PCS}. Without the well-designed distortion-aware components, like Deformable Patch Embedding (DPE) and Deformable MLP (DMLP)~\cite{zhang2022bending}, our DPPASS-T and DPPASS-S using the unified Segformer outperform the SOTA segmentation method, Trans4PASS-T, and Trans4PASS-S by 2.12$\%$ and 1.06$\%$, respectively. Meanwhile, compared with the SOTA UDA segmentation methods, our DPPASS also yields the best performance. Specifically, our DPPASS achieves 1.47$\%$ and 2.45$\%$ mIoU increment with Segformer-B1 and Segformer-B2 backbones than the UDA method PCS~\cite{PCS}. This indicates our DPPASS better tackles two types of domain gaps (\ie, inherent and format gaps) between 360$^\circ$ and pinhole image domain.

Fig.~\ref{360results} shows the qualitative comparison with the supervised Segformer-1~\cite{segformer}, Trans4PASS~\cite{zhang2022bending} and our DM-PASS. For panoramic images, the larger objects have more complex distortion than the smaller ones, thus it is more difficult to completely and neatly segment these large objects, such as sidewalks, walls, etc. Obviously, our DPPASS has \textit{significantly better} segmentation results on these larger objects with greater distortion, as shown in Fig.~\ref{360results}. The quantitative results in Tab.~\ref{perclass} also show that our DPPASS-S outperforms the Trans4PASS-S by 6.56$\%$ IoU on the sidewalk class. For the classes for autonomous driving, such as persons, riders, and motorcycles, the white dotted boxes in Fig.~\ref{360results} show that our DPPASS-S achieves much better segmentation results than the Trans4PASS-S. The IoU increments of person, rider, and motorcycle categories are +7.14$\%$, +5.74$\%$, and +9.57$\%$, respectively.

We also evaluate our DM-PASS on the WildPASS test set. Tab.~\ref{wildpass} shows the quantitative results, where our DPPASS approach consistently outperforms the SOTA UDA segmentation method Trans4PASS~\cite{zhang2022bending}. The mIoU improvements of our DPPASS-T and DPPASS-S over the Trans4PASS-T and Trans4PASS-S are 5.71$\%$ and 0.62$\%$, respectively. It is worth noting that our DPPASS achieves a dramatic increase of mIoU by 5.71$\%$ higher than the SOTA method Trans4PASS (60.38$\%$ vs. 54.67$\%$) with the Segformor-B1 backbone. This validates the problem that naively treating the ERP and pinhole images equally leads to less optimal UDA segmentation performance. Therefore, the two types of domain gaps between the 360$^\circ$ and pinhole image domain, defined by our work, are pivotal.

\section{Ablation study and Analysis}
\noindent{\textbf{Dual Projection vs. Single Projection.}}
The dual-projection training means ERP and tangent images are individually used in dual paths while the single projection only uses the ERP in both paths. So the results of dual projection (49.53$\%$) and single projection (45.22$\%$) show the superiority of dual-projection training. This also indicates the tangent projection can bring complementary benefits to the standard domain alignment.

\noindent{\textbf{Rationality of Prediction Consistency Training}}
The tangent projection of the same sphere data utilized in our work can be formulated as a data augment operation. In classic consistency training procedure, data augment methods are always leveraged before and after the forward propagation process. In our work, we use the tangent projection before and after the forward propagation to make our model aware of the distortion variation. Numerically, with our prediction consistency training, +7.13$\%$ mIoU improvement is obtained than the source pretrained model.

\begin{table}[]
\centering
\renewcommand{\tabcolsep}{10pt}
\resizebox{0.35\textwidth}{!}{
\begin{tabular}{cccccl}
\toprule
\multicolumn{4}{c}{Losses}                              & \multirow{2}{*}{mIoU} & \multirow{2}{*}{$\bigtriangleup$} \\ \cmidrule{1-4}
$\mathcal{L}_s$ & $\mathcal{L}_g$  & $\mathcal{L}_{pc}$ & $\mathcal{L}_{fc}$ & &                                        \\ \midrule
\checkmark & & & &42.40 & -                                       \\ \midrule
\checkmark & \checkmark & & & 50.12 & +7.72                           \\ \midrule
\checkmark & & \checkmark & & 49.53 &  +7.13                            \\ \midrule
\checkmark &  &  & \checkmark & 47.30 & +4.90                            \\ \midrule
\checkmark & \checkmark & \checkmark & \checkmark & 55.30 & +12.9                  \\ \bottomrule
\end{tabular}}
\caption{Ablation study of different loss combinations with DS-PASS-T framework on the DensePASS test set.}
\label{lossab}
\end{table}

\begin{table}[]
\resizebox{0.47\textwidth}{!}{
\begin{tabular}{cccccc}
\toprule
\multicolumn{6}{c}{Tangent Projection} \\  \midrule
Size  & 96 $\times$ 96  & 144 $\times$ 144  & 224 $\times$ 224  & 384 $\times$ 384  & 512 $\times$ 512  \\ \midrule
mIoU  &49.98 & 52.22 &\textbf{55.30} &55.17 &52.56      \\ \bottomrule
\end{tabular}}
\caption{Ablation study of different Tangent-Projection size with DS-PASS-T framework on the DensePASS test set.}
\label{lossts}
\end{table}

\noindent{\textbf{Rationality of Feature Contrastive Training}}
\begin{figure}
    \centering
    \includegraphics[width=0.49\textwidth]{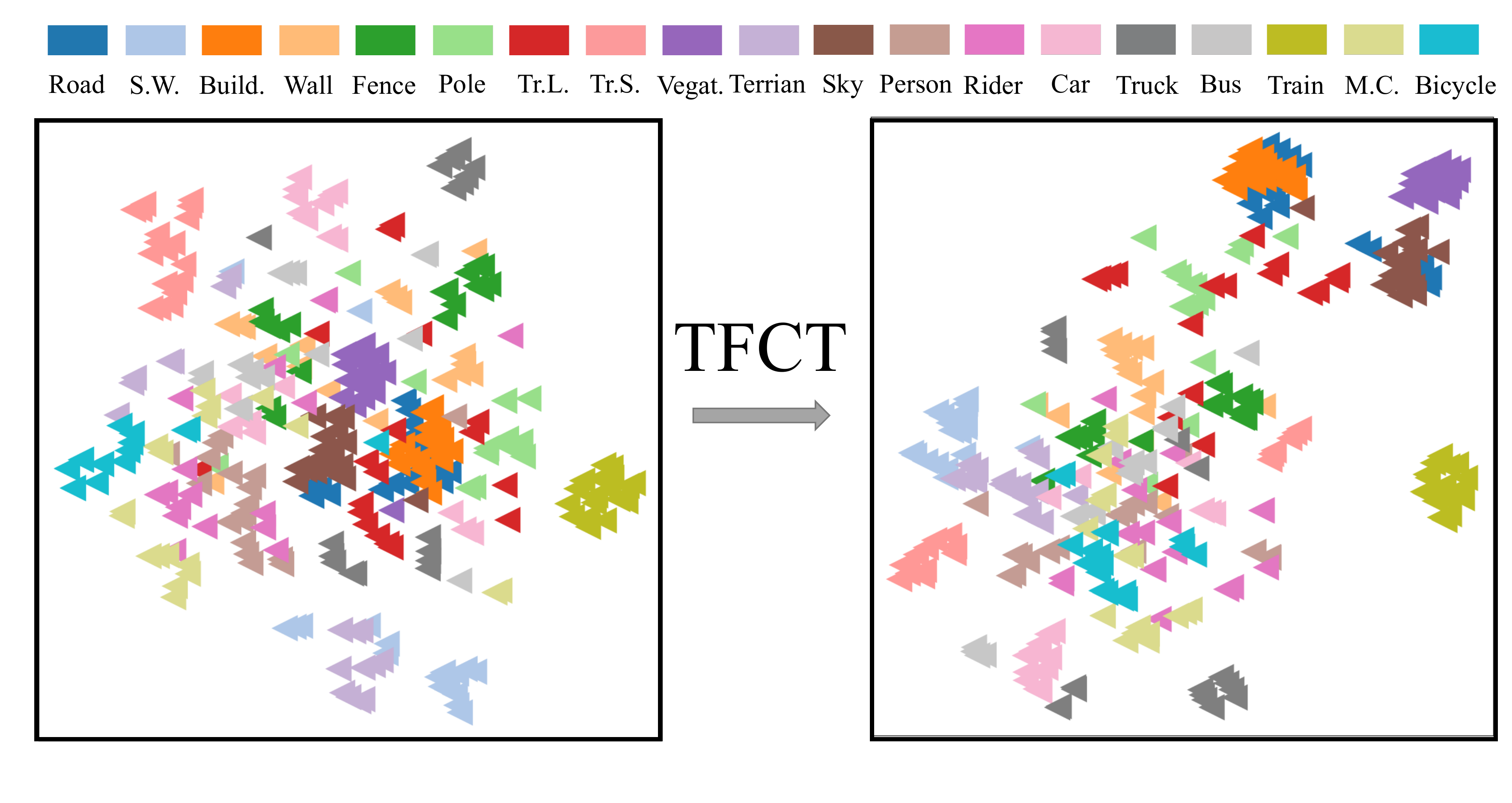}
    \caption{TSNE visualization of features with and without our proposed TFCT module.}
    \vspace{-12pt}
    \label{TFCT_tsne}
\end{figure}
Since the tangent projection $f_{E2T}(\cdot)$ is performed on the ERP input and the ERP features equally without shuffling the location, the two sequences of features are in one-to-one correspondence. Intuitively, we leverage this correspondence to distinguish the distortion information. As shown in Tab.~\ref{lossab}, the feature-wise contrastive learning give +4.90$\%$ mIoU increment compared with the supervised baseline. Qualitatively, the feature embeddings are shown in Fig.~\ref{TFCT_tsne}. As our TFCT acts on the representation space and provides feature-wise alignment, the features are pushed closer to each category against distortion and object deformation.

\noindent{\textbf{Rationality of Intra-Projection Training.}}
Both ERP and TP images provide critical domain knowledge, including style and distortion features. Intuitively, we propose to utilize this critical information in both projections to facilitate knowledge transfer. The adversarial module in each path makes the models learn complementary domain knowledge in different scales (whole ERP $\&$ Tangent Patch). Numerically, as shown in Tab.~\ref{lossab}, the Intra-Projection loss $\mathcal{L}_g$ achieves +7.72$\%$ mIoU increment than the baseline.

\noindent{\textbf{Loss functions.}} We conduct ablation experiments on the DensePASS dataset to analyze the impact of the supervised loss $\mathcal{L}_s$, the intra-porjection loss $\mathcal{L}_g$ (Eq.~\ref{gan4}), the prediction consistency training loss $\mathcal{L}_{pc}$ (Eq.~\ref{loss_con}) and the TFCT loss $\mathcal{L}_{fc}$ (Eq.~\ref{loss_tfct}) in DPPASS. In Tab.~\ref{lossab}, different combinations of losses are applied. It is obvious that the intra-porjection loss $\mathcal{L}_g$ reduces the inherent gaps which are caused by the different sensors, and brings an increase of 4.93$\%$ in the mIoU. For the cross-model modules, we can see that the prediction consistency training loss $\mathcal{L}_{pc}$ which imposes the prediction consistency between different representations of the same spherical data gives an improvement of 7.13$\%$ in mIoU. As for the TFCT $\mathcal{L}_{fc}$, it also contributes positively to the decrease of the large domain gaps between the 360$^\circ$ image domain and the pinhole image domain well with 4.90$\%$ mIoU improvement over supervised baseline.

\noindent{\textbf{TP patch size}}. Tab.~\ref{lossts} reports the mIoU($\%$) with different TP patch sizes on the DensePASS dataset. It shows that the optimal patch size is 224 $\times$ 224. Too large or too small projection size impedes the segmentation performance, the best trade-off patch size is 224 $\times$ 224.

\noindent \textbf{Trade-off ratio of \textbf{$\alpha$ and $\beta$}}. Tab.~\ref{abparam} reports the mIoU($\%$) of our DPPASS-T with different ratios of $\alpha$ and $\beta$ on the DensePASS test set. The best trade-off weights for $\mathcal{L}_{pc}$ and $\mathcal{L}_{fc}$ are $\alpha$ = 0.02 $\beta$ = 50.

\noindent{\textbf{Inference cost}}. For one ERP with the size of 400 $\times$ 2048, the inference costs of our DPPASS-S and the Trans4PASS-S are 108.84G and 251.08G in FLOPs, respectively. Our DPPASS reduce 60$\%$ computational cost while achieving +1$\%$ increment than the SOTA Trans4PASS method.

\begin{table}[t!]
    \centering
    \renewcommand{\tabcolsep}{8pt}
    \small
    \begin{tabular}{cccccc}
    \toprule
    $\alpha$&0.001&0.01&0.02&0.05&0.1\\
    \midrule
    mIoU&48.25&50.01&\textbf{50.12}&48.12&47.44\\
    \midrule
    \midrule
    $\beta$&10&20&50&100&200\\
    \midrule
    mIoU&46.97&47.21&\textbf{49.53}&49.38&42.1\\
    \bottomrule
    \end{tabular}
    \caption{Ablation study of hyper-parameters $\alpha$ and $\beta$. The reported results are trained with the combination of the supervised loss and the loss terms $L_{pc}$ and $L_{fc}$ based on our DPPASS-T.}
    \vspace{-6pt}
    \label{abparam}
\end{table}

\section{Conclusion}
In this paper, we studied a new problem by refining the domain gaps between the panoramic and pinhole images into two types: the inherent gap and the format gap. We accordingly proposed DPPASS, the first dual-projection UDA framework, taking ERP and tangent images as input to each path to reduce the domain gaps.
We introduced intra-projection training to reduce the inherent gap while the format gap was addressed by the cross-projection training. Importantly, the TP path can be removed after training, adding no extra inference cost.
Our DPPASS significantly surpassed the prior UDA methods for panoramic image semantic segmentation and achieved new SOTA performance.


\section*{Acknowledgements}
This work was supported by the National Natural Science Foundation of China (NSF) under Grant No. NSFC22FYT45. 

\clearpage
{\small
\bibliographystyle{ieee_fullname}
\bibliography{egbib}
}

\end{document}